\documentclass[letterpaper]{article} 
\usepackage{aaai25}  
\nocopyright
\usepackage{times}  
\usepackage{helvet}  
\usepackage{courier}  
\usepackage[hyphens]{url}  
\usepackage{graphicx} 
\usepackage{natbib}  
\usepackage{caption} 
\usepackage{algorithm}
\usepackage[noend]{algpseudocode}
\usepackage{bm}
\usepackage{amsmath}
\usepackage{amssymb}
\usepackage{booktabs} 
\usepackage{multirow} 
\usepackage{caption}
\usepackage{subcaption}

\usepackage{newfloat}
\usepackage{listings}
\DeclareCaptionStyle{ruled}{labelfont=normalfont,labelsep=colon,strut=off} 
\floatstyle{ruled}
\newfloat{listing}{tb}{lst}{}
\floatname{listing}{Listing}
\usepackage[inline]{enumitem}

\usepackage{todonotes}

\title{
Optimizing Vital Sign Monitoring in Resource-Constrained Maternal Care:\\ An RL-Based Restless Bandit Approach}
\author {
    Niclas Boehmer\textsuperscript{\rm 1},
    Yunfan Zhao\textsuperscript{\rm 1},
    Guojun Xiong\textsuperscript{\rm 1}, 
    Paula Rodriguez-Diaz\textsuperscript{\rm 1},\\ 
    Paola Del Cueto Cibrian\textsuperscript{\rm 2},
    Joseph Ngonzi\textsuperscript{\rm 3},
    Adeline Boatin\textsuperscript{\rm 2},
    Milind Tambe\textsuperscript{\rm 1}
}
\affiliations {
    \textsuperscript{\rm 1}School of Engineering and Applied Sciences, Harvard University, USA\\
    \textsuperscript{\rm 2}Department of Obstetrics and Gynecology, Massachusetts General
Hospital, Harvard Medical School,  USA\\
    \textsuperscript{\rm 3}Mbarara University of Science and Technology, Uganda\\ 
nboehmer@g.harvard.edu, yunfanzhao@fas.harvard.edu, xionggj1@gmail.com ,
prodriguezdiaz@g.harvard.edu,
pdelcueto@mgh.harvard.edu,
jngonzi@must.ac.ug, 
adeline\_boatin@mgh.harvard.edu,
milind\_tambe@harvard.edu
}

\usepackage{siunitx}
\usepackage{bibentry}
\usepackage{cleveref}
\usepackage{nicefrac}

\newcommand{\sname}[1]{$\mathtt{#1}$}

\begin{document}

\maketitle

\begin{abstract}
    Maternal mortality remains a significant global public health challenge. One promising approach to reducing maternal deaths occurring during facility-based childbirth is through early warning systems, which require the consistent monitoring of mothers' vital signs after giving birth. Wireless vital sign monitoring devices offer a labor-efficient solution for continuous monitoring, but their scarcity raises the critical question of how to allocate them most effectively. We devise an allocation algorithm for this problem by modeling it as a variant of the popular Restless Multi-Armed Bandit (RMAB) paradigm. In doing so, we identify and address novel, previously unstudied constraints unique to this domain, which render previous approaches for RMABs unsuitable and significantly increase the complexity of the learning and planning problem. To overcome these challenges, we adopt the popular Proximal Policy Optimization (PPO) algorithm from reinforcement learning to learn an allocation policy by training a policy and value function network. We demonstrate in simulations that our approach outperforms the best heuristic baseline by up to a factor of $4$.  
\end{abstract}

\section{Introduction} \label{intro}
Each year, more than $250,000$ women lose their lives during and following pregnancy and childbirth \cite{who_maternal_mortality_2024}, with the first 24 hours post-delivery being particularly perilous \cite{Li1996, Dol2022}. 
A significant contributing factor to this tragic statistic is the poor quality of care available in under-resourced communities \cite{CrearPerry2021}. 
Consequently, there is growing interest in different ways of improving peripartum care to prevent life-threatening complications such as hemorrhage, hypertensive disorders, and sepsis.
One key approach is through the monitoring of maternal vital signs, which can be used to identify complications early on via early warning systems that provide an opportunity for timely clinical intervention \cite{vousden2019effect}.

In fact, the \citet{world2016standards} recommends close monitoring of maternal vital signs in the first 24 hours after birth, thereby highlighting the importance of maternal vital signs in high-quality maternal care. 
Traditionally, this monitoring is conducted by healthcare providers who manually measure vital signs at regular intervals. 
However, even in well-resourced settings, the recommended monitoring frequency poses a substantial burden. In resource-limited settings, meeting these guidelines has been very difficult to achieve \cite{mugyenyi2021quality,semrau2017outcomes}.

\begin{figure}[t!]
\centering
\includegraphics[width=3.4cm]{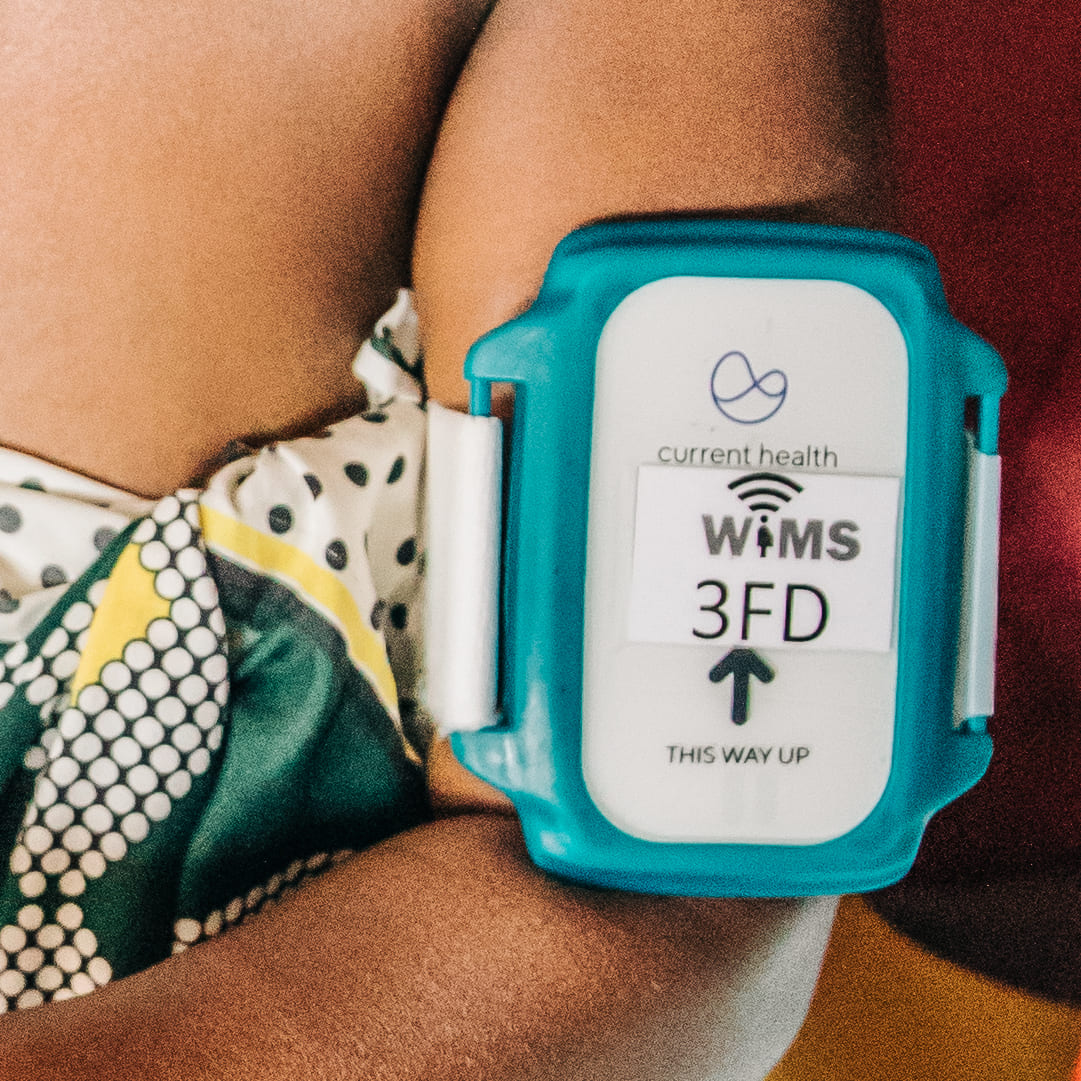}
\caption{Wireless vital sign monitoring device on the arm of a mother.} \vspace{-0.5cm}\label{fig:device}
\end{figure}

An automated alternative to measuring vital signs is the use of wearable or wireless vital sign monitoring devices, often in the form of wireless biosensors; see \Cref{fig:device} for a picture of such a device \cite{boatin2016wireless, boatin2023implementation}. 
These sensors continuously measure and transmit the mother's vital signs, with the option to trigger alerts if abnormalities in vital signs are detected. 
These alerts provide the opportunity for clinicians to initiate appropriate medical responses in real time when needed. 
Thus, automated vital sign monitoring using wireless biosensors provides an opportunity to implement early warning systems in resource-constrained environments, where to date, human resource constraints have limited the ability to check vital signs consistently. Such systems have been demonstrated to be functional and acceptable in these settings \cite{boatin2016wireless, ngonzi2017functionality}.
However, in practice, while these monitoring devices enhance human resources, their availability will still be severely limited, leading to the central research question addressed in this paper: 
\emph{Who should wear a monitoring device and for how long?}

While our research is motivated by several hospital settings in Uganda and Ghana, the specific motivation for this study 
comes from the Department of Obstetrics and Gynecology at the Mbarara Regional Referral Hospital in Mbarara,  Uganda.
With approximately \num{9000} deliveries annually, this hospital serves as the primary referral center for southwestern Uganda. It has been piloting the use of wireless vital sign monitors during the critical 24 hour period after birth \cite{boatin2021wireless,boatin2023implementation,mugyenyi2021quality}.

We tackle the problem of allocating monitoring devices by framing it as an instance of the popular restless multi-armed bandit (RMAB) model. This model has been successfully applied to distribute limited resources in different healthcare contexts  \cite{ayer2019prioritizing} as well as in areas beyond healthcare, such as anti-poaching \cite{qian2016restless} and machine maintenance \cite{abbou2019group}.
However, unique challenges arise in our application, rendering existing solution methods inadequate.
Specifically, in contrast to classic RMAB problems, our allocation setting places several novel constraints on the allocation, for instance, each mother must be allocated a device for a minimum and maximum duration, and once a device is removed from a mother, it cannot be (easily) reassigned to her at a later point (see \Cref{sec:Problem}). 
Furthermore, currently, no historical data or other relevant features of arriving mothers are available to the algorithm in the intended deployment setting. Consequently, the decision of when to remove a device after the minimum monitoring period must be based solely on the mother's  so-far recorded vital signs. Therefore, our algorithm learns in an online manner, leveraging information collected from previously monitored mothers to make decisions for new ones.
To address these challenges, we adopt the Proximal Policy Optimization (PPO) algorithm, a reinforcement learning technique that has proven effective in diverse domains \cite{schulman2017proximal} such as video games \cite{kristensen2020strategies}, robotics \cite{melo2021learning}, and autonomous vehicles \cite{guan2020centralized}. Our approach leverages PPO's strengths in learning robust policies under complex constraints and dynamic environments.
In sum, our main contributions are: 
\begin{itemize}
\item We are the first to identify and formalize the algorithmic problem of allocating scarce wireless vital sign monitoring devices, with novel real-world constraints previously unstudied by the resource allocation literature.
\item We develop a modern reinforcement learning-based algorithm to learn and make allocation decisions in real time using RMABs, contributing to both the application domain and extending the RMAB literature.
\item We demonstrate in simulation that our algorithm significantly outperforms natural heuristic baselines, achieving improvements ranging from $100\%$ to $400\%$, which gives a good indication of the usefulness of AI approaches for this problem. Moreover, we conduct a first analysis on limited data from the Mbarara Regional Referral Hospital and describe the next steps toward responsible real-world AI deployment, including additional data collection, quality control, ethics reviews, and field trials.
\end{itemize}

\paragraph{Related Work on RMABs}\label{sec:RW}

There is a rich body of work on devising allocation algorithms for scarce resources using the restless bandits model. 
Different variants of this model have been studied, differing in the information available to the planner \cite{chen2024contextual}, the constraints on budget usage \cite{rodriguez2023flexible,dexun2022efficient}, and resource allocation strategies \cite{mao2024time}. 
The work of \citet{DBLP:conf/atal/MateBSGT22} studies streaming restless multi-armed bandits, where arms appear and disappear over time, which is also the case in our problem. 
The primary distinction between their setting and ours lies in their assumption that arms have discrete states with known transition dynamics, whereas in our problem, transition dynamics are unknown and states are not restricted to being discrete. Moreover, we have additional constraints placed on the allocation.
In contrast, the work of \citet{DBLP:conf/atal/ZhaoB0ZNTTT24} proposes a reinforcement learning-based solution for streaming bandits where transition dynamics are unknown and states consist of a single continuous value. 
However, their method is not applicable to our setting, as it cannot accommodate the additional allocation constraints specific to our problem (specifically, the lambda network they use is incompatible with constraints beyond the standard budget one). Moreover, their approach relies on feature information that is unavailable in our context and is limited to simpler state spaces, which do not adequately capture the complexity of vital sign profiles (see \Cref{sec:Problem}). \looseness=-1

\section{Application, Modeling, and  Challenges} \label{sec:Problem}

\subsection{Application Details} \label{sec:appl}

We outline the characteristics of the problem encountered in our application domain. Our setting is the  maternity unit of the Mbarara Regional Referral Hospital \cite{boatin2021wireless,boatin2023implementation}:
Mothers arrive at the maternity unit and deliver throughout the day. 
After delivery, mothers remain in the maternity unit for some time before being redirected to other care measures or discharged. 
Mothers will wear a monitoring device during some time they spend in the unit.

Wearing a monitoring device has no direct impact on the mother’s vital signs or health. However, there is a clear indirect impact: If a monitored vital sign deviates from the preset normal range, an alert is sent to the responsible clinician’s phone. While some alerts may be disregarded due to capacity constraints or other factors, in most cases, the clinician will visit the patient, manually assess any abnormal vital signs, and initiate appropriate clinical interventions if needed. These interventions are expected to positively influence the mother’s health and stabilize her vital signs.

There are several external constraints imposed on the allocation of monitoring devices. First, every mother should wear the device for a minimum duration, for instance, the initial two hours after birth, which are particularly high-risk. Second, once a device is removed from a mother, it cannot be (easily) reassigned to her, as she will transition to a different set of care protocols. Third, each mother is only eligible to wear the device for the first $24$ hours after birth, as this is the targeted monitoring period for the program.
No feature information about the mother—such as demographic details or historical medical data—is available to the algorithm, constituting a safeguard for data privacy.\footnote{We note that in some hospitals such information exists on paper and could be digitalized if needed. However, an algorithm that does not require features and historical data is naturally much easier to deploy and preferable from a privacy and safety perspective.}
Consequently, the decision when to remove the device from a mother after her minimum monitoring period has ended must be made purely based on her previously recorded vital signs.

The objective of the allocation strategy is to minimize occurring complications and maintain the vital signs of all patients—whether they are wearing a device or not—within the normal range during their stay in the maternity unit. This implies that patients at higher risk of developing abnormal vital signs indicating potential complications should be prioritized: For them, a monitoring device will trigger alerts and prompt the needed timely clinical assistance.

\subsection{Formal Modeling \& Challenges}

We model the problem of allocating monitoring devices using the popular RMAB framework. 
An instance of our problem consists of a planning horizon $T$, a budget $B$ (the number of available devices), a discount factor $\gamma$, a minimum $t_{\mathrm{min}}$ and maximum $t_{\mathrm{max}}$ number of steps a mother should be monitored, and a set $N$ of mothers (from now on called \emph{arms}). 
Each arm $i\in N$ follows a Markov Decision Process $(\mathcal{S}_i,\mathcal{A}_i=\{0,1\}, \mathbf{\Gamma}_i, R_i)$.
$\mathcal{S}_i$ represents the possible \emph{states} of arm $i$. In our application, states are multidimensional and continuous and include the current values of the vital signs, along with potentially aggregated statistics like the variability of each vital sign over recent time steps.
$\mathcal{A}_i=\{0,1\}$ represents the actions, where $0$ denotes the \emph{passive} action and $1$ denotes the \emph{active} action, i.e., allocate a device to the mother.
$\mathbf{\Gamma}_i$ describes the parameters characterizing how arm $i$'s state evolves from one step to the next conditioned on the taken action.\footnote{For example, transitions of continuous states might follow multivariate Gaussian distributions (see  \Cref{setup}), with separate distributions for the active and passive action. Then, $\mathbf{\Gamma}_i$ contains the distributions' mean and covariance matrix. In many states, applying the active action (i.e., allocating a device) will not alter the transition dynamics unless an alert is triggered. Nonetheless, it is beneficial to continue monitoring such arms as they may transition into critical states later on where alerts are generated and the transition dynamics are impacted by the active action.}
$R_i:\mathcal{S}_i\to \mathbb{R}$ is the reward function of arm $i$, penalizing states where vital signs fall significantly outside the normal range, indicating potential complications.
Additionally, each mother has an arrival $\alpha_i\in [T]$ and departure time $\beta_i\in [T]$. 
We assume that the state space $\mathcal{S}$ and reward function  $R$ are the same for each arm and are known. 
In contrast, $\Gamma_i$ is arm-specific and unknown. Additionally, $\alpha_i$ and $\beta_i$ are also arm-specific and are revealed at the corresponding timestep.  At each timestep $t\in [T]$, an arm $i$ is \emph{present} if $t\in [\alpha_i,\beta_i]$.  Let $N_t$ be the set of arms present at time $t$.  
The goal is to learn a policy $\pi$ that maps the set of currently present arms and their current states $\mathbf{s}\in \mathcal{S}^{|N_t|}$ to an action vector $\mathbf{a}\in \{0,1\}^{|N_t|}$, such that for each $t\in [T]$ and $i\in N_t$:
\begin{enumerate}
    \item $\sum_{j\in N_t} a_j\leq B$, 
    \item $a_i=1$ if $t\in [\alpha_i,\alpha_i+t_{\mathrm{min}}-1]$ ,
    \item $a_i=0$ if $t>\beta_i$ or $t\geq\alpha_i+t_{\mathrm{max}}$, and 
    \item $a_i=0$ if there is some step $t'\in [\alpha_i,t]$ in which $i$ was assigned the passive action.
\end{enumerate}
The goal is to find such a policy that maximizes the accumulated discounted reward: 
$\sum_{t\in [T]} \gamma^{t-1} \mathbb{E}_{\mathbf{s}\sim(N,\pi)} \sum_{i\in N_t} R(s_i).$
Note that due to the allocation constraints,  allocation decisions only need to be made when a new arm arrives. At that moment, the algorithm must assign the active action to the new arm (due to the minimum monitoring period). The ``only'' decision the algorithm needs to make is which arm should be flipped from the active to the passive action, i.e., from which mother we take the monitoring device needed for the new mother. 

\paragraph{Novel Challenges}
Our problem introduces three novel aspects that set it apart from existing work in restless bandits:
\begin{itemize}
    \item The standard assumption in the restless bandit literature is that states are few and discrete, which simplifies transition dynamics \cite{ninomora2023}. However, vital signs evolve in complex, continuous ways, prohibiting the discretization of the state space. Further, patient's states are characterized by multiple continuous vital sign values plus statistics about their trajectory. 
    \item Existing works on RMABs with unknown transition probabilities typically rely on arm's feature information to learn their dynamics. In our context, no such features are available, forcing the algorithm to estimate an arm’s future behavior based solely on its current state.
    \item To our knowledge, allocation constraints $2$-$4$ from above are important for many monitoring applications but have not been previously explored in the RMAB literature. 
\end{itemize}

\section{Methodology}

To address the novel challenges arising in our application domain, we employ an actor-critic approach using Proximal Policy Optimization (PPO) 
for policy updates, which is widely used in reinforcement learning.
Our algorithm requires access to a simulator ``Simulate$(i,s_i,a_i)$'' of the environment that takes as input an arm, its current state, and its action and outputs the new state of the arm.
The idea is to train a policy using \Cref{alg:main_algorithm_training} which has access to the simulator and then deploy the learned policy in the real-world, which ensures the required high-quality decision-making from the start \cite{DBLP:conf/atal/ZhaoB0ZNTTT24}. 

The actor is a policy neural network that takes as input the current state of an arm and outputs the action probability for both possible actions.
We let $\theta(a\mid s)$ denote the action probability for action $a\in \{0,1\}$ returned by the network on input $s\in \mathcal{S}$.
We act on the arms with the highest probability for the active action. Thus, the output of this network determines which arms are assigned the active action.
The critic is another neural network that takes as input the state $s\in \mathcal{S}$ of an arm and outputs the baseline estimate $V(s)$, which is the expected total discounted reward generated by this arm starting from state $s$ assuming that actions are taken following the action probabilities returned by the policy network. 
The critic network is used for updating the policy network via the PPO algorithm.
Importantly, both networks are shared among all arms, enabling arms to learn from each other—this is crucial because each arm remains in the system for only a limited time.

\Cref{alg:main_algorithm_training} proceeds in multiple epochs. 
For training purposes, each epoch deals with a separate set of arms.
Within each epoch, a fixed policy is used to make the allocation decisions while respecting all constraints. 
The behaviors of all arms are observed and stored in a buffer.
At the end of each epoch, we use the buffer to update the policy and critic networks, thereby refining the policy. 

\begin{algorithm}[t]
\caption{RL-based allocation algorithm}
\begin{algorithmic}[1]
\State Input: $n_{\mathrm{epoch}}$ instances of our problem. 
\State Initialize actor $\theta$ and  critic $\phi$

\For{each of the $n_{\mathrm{epoch}}$ instances}
    \For{$t=1,\dots, T$}\label{l:5}
        \State Let $N_t^{\mathrm{new}}:=\{i\in N_t\mid \alpha_i+t_{\mathrm{min}}< t\}$ \label{l:6}
        \State Assign the active action to all arms from $N_t^{\mathrm{new}}$ \label{l:7}
        \State Let $N_t^{\mathrm{eligible}} $ be the set of all arms $i\in N_t$ with $\alpha_i+t_{\mathrm{max}}\leq t$ and which have never been assigned the passive action since it arrived in step $\alpha_i$ \label{l:8}
        \State Use policy network to compute action probability $\theta(a_i \mid  s_i)$ for each arm $i\in N_t$ \label{l:9}
        \State Assign the active action to the $B-|N_t^{\mathrm{new}}|$ arms from $N_t^{\mathrm{eligible}}\setminus N_t^{\mathrm{new}}$ with the highest active action probability $\theta(1\mid \cdot)$ \label{l:10} 
        \For{$i\in N_t$}
        \State $s'_i= \text{Simulate}(i,s_i,  a_i)$ \label{l:11}
        \State Add tuple $(s_i,  a_i, R(s_i), s'_i)$  to buffer  \label{l:12}
        \State $s_i\leftarrow s'_i$\label{l:13}
        \EndFor
    \EndFor
    \State Update actor-critic $(\theta,\phi)$ pair via PPO using~buffer  \label{l:14}
\EndFor 
\end{algorithmic}
\label{alg:main_algorithm_training}
\end{algorithm}

Breaking down \Cref{alg:main_algorithm_training}, in Lines \ref{l:6}--\ref{l:10}, the algorithm assigns actions to all present arms: 
In Line \ref{l:7},  the algorithm assigns the active action to all arms that have been in the system for less than $t_{\mathrm{min}}$ steps, as they have not been monitored for the required minimum time. 
In Line \ref{l:10}, the remaining active actions are assigned to the arms eligible for receiving an active action in this step with the highest action probability returned by the policy network.
Then, in Line \ref{l:11}, we simulate the next state of each arm conditioned on the assigned action and update its state accordingly~in~Line~\ref{l:13}.

The policy and critic networks are updated following the principles of the PPO algorithm \cite{schulman2017proximal}. Specifically, let $V(s)^j$ be the values returned by the critic network at the end of epoch $j$. We compute the advantage function $A^j(s,a)$ for epoch $j$,  $s\in \mathcal{S}$ and $a\in \{0,1\}$, which quantifies the benefit of taking a certain action $a$ in state $s$ compared to the current policy as 
$A^j\left(s, a\right)=Q^j\left(s, a\right)-V^j\left(a\right)$,
where $Q^j(s,a)$ is the expected discounted cumulative reward for the completion of the current episode under the current policy for an arm in state $s$ to which action $a$ is applied in this step. 
The advantage function is then incorporated into the actor’s policy gradient to update the policy network in the actor, following the standard PPO procedure.

When running the algorithm in testing, we execute Lines \ref{l:5} to \ref{l:13} with the trained policy network. 

\iffalse

We adapt PPO \cite{schulman2017proximal}, a policy gradient algorithm with an actor-critic structure. 

In line 14, to update the actor-critic pair, we first update the Q-function as follows:

\begin{align*}
Q^t(s_n,a_n)\leftarrow &(1-\alpha_q) Q^{t-1}(s_n,a_n)  \\
&+\alpha_q \left(R(s_n), \gamma \max_a Q^{t-1}(s_n',a)\right),
\end{align*}

where $\alpha_q$ is the learning rate. After performing an update of the Q-function in the critic, we compute the advantage estimate: 

$$
A\left(\mathbf{s}, \mathbf{a}\right)=Q\left(\mathbf{s}, \mathbf{a}\right)-V\left(\mathbf{s}\right).
$$ 

where the values $A(s,a), Q(s,a), V(s)$ are evaluated based on the current policy $\pi$. Intuitively, the advantage estimate tells us how much better $\mathbf{a}$ is compared to the current policy $\pi$. The advantage estimate is then plugged into the policy gradient to update the actor.

\fi

\section{Experiments}\label{exp}
We present our experiments using data from a publicly accessible de-identified vital sign database, which offers rich and high-quality data for conducting comprehensive experiments. The goal of this section is to demonstrate the general capabilities of our algorithm to distribute vital sign monitoring devices; we revisit our initial use case of maternal care in \Cref{uganda}. All experiments are conducted in simulation; our algorithm is only applied to simulated vital sign profiles.
In \Cref{setup}, we describe our setup, including the datasets used, the instance generation, the trained simulator, and the baselines employed. In \Cref{results}, we present and analyze our results. 

\subsection{Setup} \label{setup}
\paragraph{Domain}
The experiments in this section are based on data from the widely used MIMIC-III \cite{johnson2016mimic} and MIMIC-IV \cite{johnson2023mimic} datasets, which have become popular and influential in computer science research. Both datasets contain de-identified clinical data from thousands of patients who stayed in critical care units at Beth Israel Deaconess Medical Center in Boston over different periods, including vital sign measurements typically recorded at one-hour intervals. In our experiments, each arriving arm corresponds to a new patient entering the critical care unit. 
For MIMIC-III, we use the vital signs
\begin{enumerate*}[label=(\roman*)]
  \item heart rate, 
  \item speed of breathing (respiratory rate), and 
  \item blood oxygen saturation (SPO2)
\end{enumerate*}, while for MIMIC-IV, we use 
\begin{enumerate*}[label=(\roman*)]
  \item heart rate, 
  \item respiratory rate, and
  \item skin temperature.
\end{enumerate*}
We normalize each vital sign between $0$ and $1$ using min-max normalization. For each patient, we take the median vital sign value at each hour. Thus, one timestep corresponds to one hour. We exclude patients with fewer than $10$ data points.

\paragraph{Simulator} 
The state representation of patients includes, for each vital sign, its current value and the variance of the value over the last five timesteps. The reward function assigns a reward of $0$ if all vital sign values fall within the normal range. 
For each abnormal vital sign, we incur a negative reward that shrinks exponentially with the extent of the deviation from the normal range.\footnote{Our definitions of the abnormality thresholds for each sign largely follow \cite{boatin2021wireless}. See \Cref{app:simul} for details.} The exponential penalties model the increasing severity associated with larger deviations from the normal range.

The patient’s behavior is governed by a multivariate Gaussian distribution defined over the vital sign values in the current step and in the next step.
We sample the initial state of a patient from this distribution by taking a sample and using only the sampled vital sign values in the current step. 
Under the passive action, the next state is sampled from the conditional distribution of the Gaussian, given the current state. 
Under the active action, we make a case distinction. If all vital signs are within the normal range, the state transitions as under the passive action since no alert is triggered, and the device does not influence the patient’s trajectory.  
If any vital sign is abnormal, with a probability of 30\%, the state transitions as under the passive action (modeling cases where clinicians do not respond to the alert, which occur approximately 30\% of the time in the study by  \citet{boatin2021wireless,boatin2023implementation}).
Otherwise, the abnormal vital signs are probabilistically adjusted towards the normal range before sampling the next state, reflecting the positive effects of clinical intervention following an alert (see \Cref{app:simul} for details). 

\subsection{Results}\label{results}
\begin{figure*}[h!]
     \begin{minipage}{\textwidth}
         \centering
         \includegraphics[width=0.6\textwidth]{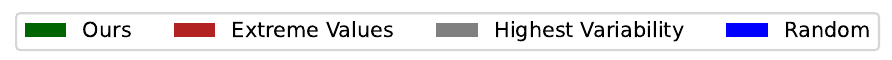}
     \end{minipage} 
     \begin{minipage}{\textwidth}
         \centering
         \includegraphics[width=1\textwidth]{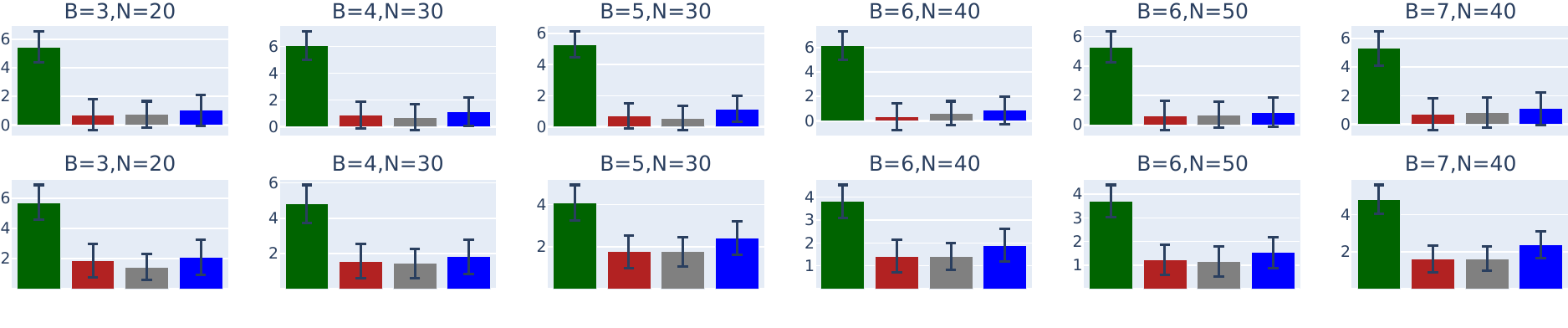}
         \vspace{-0.9cm}
     \end{minipage}
        \caption{Results on MIMIC-III (top) and MIMIC-IV (bottom), averaged over $100$ random seeds for varying budget $B$ and number of patients $N$. The error bars show the standard error of the generated reward, which is normalized by subtracting the reward of the \sname{No \ Action} baseline and then dividing by $N$. See Table~\ref{table:appendix_main_res} in Appendix~\ref{sec:appendix_exps} for additional experimental results.} 
        \label{fig:results}
        \vspace{-0.4cm}
\end{figure*}

\paragraph{Instances}
We set $T=100$, $t_{\mathrm{min}}=3$, $t_{\mathrm{max}}=25$, and assume that patients leave $50$ steps after they join. We vary the budget $B$.
Initially, there are $B$ patients, and every five steps new patients join. 
We report the number of patients $N$, which describes the ``typical'' number of patients in the system.
The inner workings of our instances are best understood by looking at a concrete example: Let us consider $B=3$ and $N=20$. Initially, there are $B=3$ patients, and every five timesteps, two (i.e., $\nicefrac{N}{10}$) new patients arrive. Since each patient leaves $50$ steps after they join and every five steps two patients arrive, the number of patients in the system gradually grows to $20$ and stabilizes at $20$. Importantly, at every point at which the algorithm makes an allocation decision, there are only few patients to pick from. In this instance, there are five: two (i.e., $\nicefrac{N}{10}$) newly arriving patients and three (i.e., $B$) existing patients who currently hold a device. 

It remains to describe how we sample patients' transition parameters. 
For this, we fit a weighted mixture of five multivariate Gaussians (i.e. five components) on the dataset, where we partition all trajectories into tuples that include the vital signs in the current and next step.
When we sample a patient, we first select a component from the Gaussian mixture based on the component's weights. This determines the initial mean and covariance of the patient. To introduce variability, we linearly combine this mean and covariance with those of another randomly selected component, using a weight uniformly sampled between $0$ and $0.15$. \looseness=-1

\paragraph{Baselines} We refer to \sname{No Action} as the policy that does not allocate any monitoring devices.
All other baselines respect the allocation constraints. 
Recall that this means that they only need to make a decision if a new patient appears. 
Then, the algorithm needs to decide from which patient currently holding a device and having already been monitored for $t_{\mathrm{min}}$ steps, we remove the device.  
\begin{itemize}
    \item \sname{Random}:
    Randomly selects a patient.
    \item \sname{Extreme\ Values}: Selects the patient with the least abnormal vital signs, i.e., the patient with the lowest summed normalized vital sign values (where SPO2 is inverted, as lower values indicate abnormality).
    \item \sname{Highest\ Variability}: Selects the least stable patient, i.e., the patient with the highest summed variance of vital signs over the past five steps.
\end{itemize}
\sname{Extreme\ Values} is particularly intuitive, as it allocates the devices to the patients generating the lowest reward.

\paragraph{Training and evaluation} For each considered setting, we average our results over $100$ seeds. 
For each seed, we do the following: 
We train our algorithm for $n_{\mathrm{epoch}}=50$ epochs. 
At the beginning of each epoch, we create a new instance by sampling a fresh set of $N$ arms. 
Subsequently, we evaluate the trained policy along with various baselines on $50$ newly generated instances and compute the average rewards. Both in testing and evaluation, patients transition according to the simulator as described in~\Cref{setup}. 

\paragraph{Results} 
See \Cref{fig:results} for an overview of our experimental results, where we vary the budget $B$ and the number of patients $N$ and see Table~\ref{table:appendix_main_res} in Appendix~\ref{sec:appendix_exps} for results from additional settings. 
We report the reward averaged over $100$ randomly generated seeds, where we normalize the reward of our algorithms and heuristics by subtracting from it the reward of the \sname{No \ Action} baseline and then dividing by $N$. Consequently, the reported values capture the benefit of the allocated monitoring devices.
We observe that our method outperforms the baselines across all examined settings. Notably, when $B=3$ and $N=20$, we outperform the second-best baseline by $433\%$ and  $173\%$ on the MIMIC-II  and MIMIC-IV datasets, respectively; when $B=6$ and $N=50$, we outperform the second-best baseline by $431\%$ and $141\%$ on the MIMIC-III and MIMIC-IV data, respectively. Interestingly, the intuitive \sname{Extreme \ Values} and \sname{Highest \ Variability} baselines perform worse than  \sname{Random}.
The fact that both baselines are insufficient highlights the complexity and intricacy of our problem and motivates the necessity for a more intricate approach like ours. \looseness=-1

In Appendix \ref{appendix_B3}, we analyze how the vital signs influence the allocation decisions made by the algorithm.
We observe that most reassignments happen when a patient's vital signs are within a medium range and have low~variability. 

\section{Maternal Care in Mbarara: Initial Results} \label{uganda}

\begin{figure}[ht]
     \begin{minipage}
     {0.48\textwidth}
         \centering
         \includegraphics[width=1.02\textwidth]{figures/legend_cropped.pdf}
     \end{minipage}
     \begin{minipage}
     {0.48\textwidth}
         \centering
         \includegraphics[width=\textwidth]{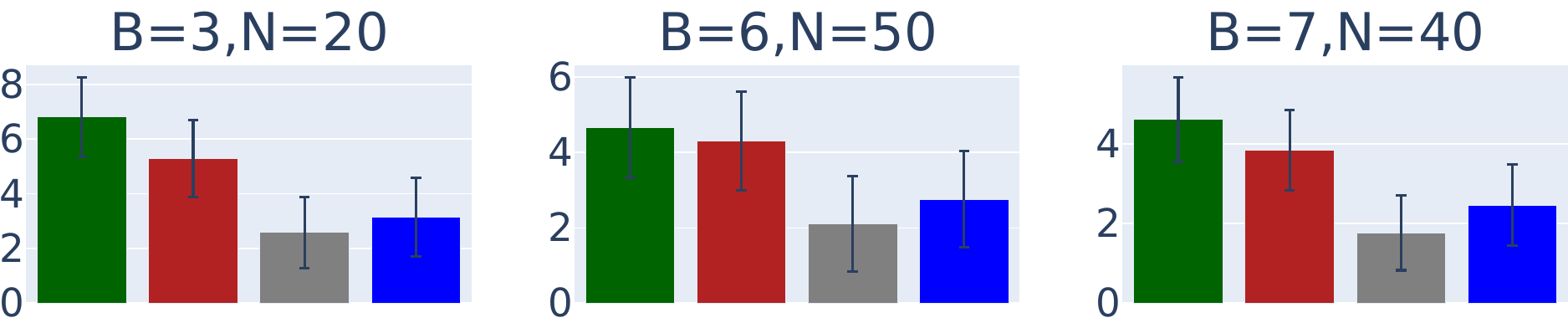}
         \vspace{-0.4cm}
     \end{minipage}
     \caption{Initial results on data from the Mbarara Hospital, averaged over $100$ random seeds. The error bars show the standard error of rewards, which are normalized by subtracting the reward of the \sname{No \ Action} baseline and then dividing by $N$ (see \Cref{sec:appendix_exps} for additional~settings).} 
     \vspace{-0.4cm}
\label{fig:results_uganda}
\end{figure}
We revisit the specific application of maternal care in Mbarara discussed in \Cref{sec:appl}.
We have access to continuous vital sign measurements from $100$ patients collected using monitoring devices at the Mbarara Regional Referral Hospital.\footnote{These trajectories were collected as part of a hybrid effectiveness-implementation trial of a wearable vital sign monitor among post-cesarean women. Women in this trial consented to wear the monitor for 24 hours. Ethics approval for this trial was obtained from the Mbarara University of Science and Technology Research Ethics Committee (17/10-18), the Uganda Council of Science and Technology (HS417ES) and the MassGeneralBrigham Institutional Regulatory Board (2019P000095). The trial was registered at clinicaltrials.gov (NCT04060667). We use the previously collected vital sign trajectories from trial participants to develop a simulator, however, vital sign data generated in these simulations do not represent and cannot be traced to real participants.}
As in \Cref{exp}, we discretize the trajectories into $60$-minute intervals by taking the median value of each vital sign recorded during the interval. The vital signs we consider are: \begin{enumerate*}[label=(\roman*)]
  \item heart rate, 
  \item respiratory rate, 
  \item and blood oxygen saturation (SPO2)
\end{enumerate*}. The rest of the setup is as~in~\Cref{exp}. 

Note that we can only fit the simulator on the 100 patients' traces available to us (we are working closely together with Mbarara University of Science and Technology to collect more data and build a more powerful simulator).
As a result,
the patients' behaviors captured by the simulator are quite simple and less stable compared to the much larger MIMIC datasets.
In the absence of complex patterns and interactions in the vital signs generated by the simulator, simpler algorithms (e.g., the random strategy) are expected to show improved behavior. 
\textit{A key question we wish to address in this section is whether the limited dataset already provides evidence that a purely random strategy is insufficient and a more involved approach is needed?}

Figure~\ref{fig:results_uganda} shows the results in three different settings. We observe that our method significantly outperforms the \sname{Random} as well as the  \sname{Highest\ Variability} strategies. Furthermore, while the difference with the second-best \sname{Extreme\ Values} is not statistically significant, there is still a trend of our algorithm showing superiority, as it outperforms this second-best baseline by $29\%$ and $20\%$ for settings $(B=3, N=20)$ and $(B=7, N=40)$, respectively.
 In light of the results from \Cref{results}, we expect these differences to grow significantly if we add additional training data for the simulator and in real-world deployment.Thus, our results provide a first evidence for the advantages of our method in the maternal care domain.

\section{Path to Deployment}\label{sec:path}
While our experiments demonstrate the potential of using RL-based algorithms for allocating monitoring devices, several important steps remain before real-world deployment. We are currently planning to collect additional vital sign trajectories in the Mbarara Regional Referral Hospital based on which we want to refine our model, especially regarding the impact of wearing a device. Once the simulator is trained on a larger and more diverse dataset, we will conduct a rigorous evaluation of the simulator and the learned policy, including assessing potential biases, verifying robustness to distribution shifts in patient populations, and making necessary adjustments. Once the policy’s decision-making is thoroughly validated, we will proceed with obtaining ethics and regulatory approval to test the policy in a real-world setting.
Recall that, as a safeguard for data privacy, no feature information about the mother is available to the algorithm.
Next, we will run a first trial in the Mbarara Regional Referral Hospital to test the implementation pipeline, safety, and acceptability of our method, and to conduct a preliminary analysis of its effectiveness. 
Assuming this study meets predefined milestones regarding feasibility, acceptability, and safety, the final phase consists of a comprehensive evaluation through a randomized controlled trial (RCT) in multiple hospitals.  
At the conclusion of a successful RCT, we will focus on the careful, responsible deployment of our system.
Throughout this entire path to deployment, we will maintain a very close collaboration with domain experts and agencies to thoroughly check for biases, and ensure steps towards a responsible deployment.
Additionally, we plan to explore the broader application of RL-based algorithms in other post-surgical care settings where monitoring devices can be used to improve patient outcomes.

\section{Conclusion}
We identified the problem of distributing wireless vital sign monitoring devices---particularly relevant in peripartum maternal care---as a novel resource allocation challenge. We introduced an RMAB-style model for this problem, which differs from previously studied models in several key aspects. Our experiments demonstrate that our RL-based allocation algorithm enables more efficient use of limited monitoring devices.
There are several promising directions for future research. The first is the path outlined in \Cref{sec:path}. Beyond this, the unique characteristics of our setting motivate the study of new variants of RMAB models. For instance, it would be interesting to develop algorithms with performance guarantees for traditional RMAB settings where arms' MDPs are discrete and known but some of our allocation constraints must be respected.
Our application also raises additional algorithmic questions. One notable challenge is optimizing the design of alerts sent by monitoring devices: While sending more alerts increases the likelihood of an alert being sent before or during a complication, it also increases the burden on clinicians and reduces their responsiveness to each individual alert. An intricate challenge for future work is determining optimal thresholds for vital sign alerts that strike the right balance between timely detection of complications and minimizing alert fatigue.

\newpage
\onecolumn
\appendix

\section{Simulator Details}\label{app:simul}

\paragraph{Normal Vital Sign Range}
To define the normal range, we primarily follow the thresholds used for alerts signaling abnormal vital sings in the study on vital sign monitoring devices for maternal health in Mbarara \cite{boatin2021wireless} featured earlier: A heart rate above $120$, a temperature above $38$C, a respiratory rate above $30$, and an SPO2 rate below $90$ are considered abnormal. Unlike \cite{boatin2021wireless}, we only use one-sided thresholds here, as our current pipeline is limited to monotonic reward functions.

\paragraph{Reward Function}
For a heart rate \( h \), the penalty is \( -\exp\left(\nicefrac{|h - 120|}{17}\right) \). For a temperature \( t \), the penalty is \( -\exp\left(\nicefrac{|t - 38.0|}{2}\right) \). For a respiratory rate \( r \), the penalty is \( -\exp\left(\nicefrac{|r - 30|}{5}\right) \). For an SPO2 \( s \), the penalty is \( -\exp\left(\nicefrac{|s - 90|}{4}\right) \).

\paragraph{Effect of Intervention}
The following describes what happens to each abnormal vital sing of a patient currently wearing a device that is examined by a doctor ($70\%$ probability).
For skin temperature, we reduce it by a sample from a normal distribution with mean $1.5$ and standard deviation $0.5$; for pulse rate, we sample from a distribution with mean $15$ and standard deviation $5$; for respiratory rate, we sample from a distribution with mean $10$ and standard deviation $3.33$. For SPO2, we increase the value by a sample from a normal distribution with mean $3$ and standard deviation $1$.

\iffalse
\section{Additional Data Description from Mbarara Regional Referral Hospital} \label{sec:appendix_mbarara}
In Section~\ref{uganda} we use continuous vital sign measurements from 100 patients, collected using monitoring devices at the Mbarara Regional Referral Hospital. These data were gathered as part of a hybrid effectiveness-implementation trial involving a wearable vital sign monitor used by post-cesarean women, who consented to wear the monitor for 24 hours. Ethics approval for this trial was obtained from the Mbarara University of Science and Technology Research Ethics Committee (17/10-18), the Uganda Council of Science and Technology (HS417ES) and the MassGeneralBrigham Institutional Regulatory Board (2019P000095). The trial was registered at clinicaltrials.gov (NCT04060667). We use the previously collected vital sign trajectories from trial participants to develop simulation models, however, vital sign data generated in these simulations do not represent and cannot be traced to real participants.
\fi

\section{Additional Experimental Results}
\label{sec:appendix_exps}
In this section, we provide additional experimental results.

\subsection{Implementation Details of \Cref{alg:main_algorithm_training}}\label{app:details}

\begin{table}[htb!]
    \centering
\scalebox{.9}{
    \begin{tabular}{@{}lc@{}}
        \toprule
        hyperparameter & value \\ 
        \midrule
        Number of hidden layers in policy network & 2 \\
        Number of neurons per hidden layer & 16 \\
        agent clip ratio  & 2\\
        start entropy coeff & 0.5\\
        end entropy coeff & 0\\
        actor learning rate & 2.0e-03\\
        critic learning rate & 2.0e-03\\
        trains per epoch & 20\\
        discount factor & 0.9\\
        \bottomrule
    \end{tabular}
    }
        \caption{Hyperparameter values.}
    \label{table:hyperparameter}
\end{table}

\subsection{Results for MIMIC Dataset}
\begin{table*}[htb!]
\scalebox{.9}{
        \begin{tabular}{c@{}p{8mm}c cccc@{} c cccc@{}}
        \toprule
        \multirow{2}{*}{\textbf{$B$}} & \multirow{2}{*}{\ \ \textbf{$N$}} & \multicolumn{4}{c}{\textbf{MIMIC-IV}} 
         & & \multicolumn{4}{c}{\textbf{MIMIC-III}}\\
        \cmidrule{3-6} \cmidrule{8-11}
    & & Ours & Random & Extreme Val & High Var
        & & Ours & Random & Extreme Val & High Var\\
        \midrule
        b = 3 & \quad 20 &  5.73 $\pm$ 1.16 & 2.1 $\pm$ 1.14 & 1.86 $\pm$ 1.11 & 1.45 $\pm$ 0.88 & &  5.44 $\pm$ 1.08 & 1.02 $\pm$ 1.1 & 0.71 $\pm$ 1.07 & 0.73 $\pm$ 0.93  \\
        b = 4 & \quad 20 &  4.2 $\pm$ 0.94 & 2.82 $\pm$ 0.93 & 2.1 $\pm$ 0.91 & 2.1 $\pm$ 0.93 & &  5.85 $\pm$ 1.46 & 1.29 $\pm$ 1.31 & 0.63 $\pm$ 1.3 & 0.57 $\pm$ 1.04 \\
        b = 5 & \quad 20 &  4.88 $\pm$ 1.13 & 3.41 $\pm$ 1.12 & 2.25 $\pm$ 1.11 & 2.52 $\pm$ 1.0 & &  6.43 $\pm$ 1.43 & 1.59 $\pm$ 1.38 & 1.04 $\pm$ 1.36 & 1.27 $\pm$ 1.09  \\
        \midrule
        b = 4 & \quad 30 &  4.82 $\pm$ 1.07 & 1.82 $\pm$ 0.98 & 1.56 $\pm$ 0.97 & 1.45 $\pm$ 0.82 & &  6.07 $\pm$ 1.04 & 1.12 $\pm$ 1.04 & 0.87 $\pm$ 1.02 & 0.7 $\pm$ 0.96  \\
        b = 5 & \quad 30 & 4.09 $\pm$ 0.83 & 2.42 $\pm$ 0.79 & 1.77 $\pm$ 0.77 & 1.77 $\pm$ 0.69 & &  5.29 $\pm$ 0.83 & 1.17 $\pm$ 0.82 & 0.71 $\pm$ 0.8 & 0.57 $\pm$ 0.77  \\
        b = 6 & \quad 30 & 5.09 $\pm$ 0.96 & 2.73 $\pm$ 0.91 & 1.73 $\pm$ 0.9 & 1.91 $\pm$ 0.8 & &  5.94 $\pm$ 1.16 & 1.38 $\pm$ 1.11 & 0.68 $\pm$ 1.09 & 0.84 $\pm$ 0.91  \\
                \midrule
        b = 5 & \quad 40 & 4.28 $\pm$ 0.73 & 1.72 $\pm$ 0.74 & 1.47 $\pm$ 0.73 & 1.33 $\pm$ 0.54 & &  5.77 $\pm$ 1.07 & 0.95 $\pm$ 1.04 & 0.5 $\pm$ 1.02 & 0.66 $\pm$ 0.94  \\
        b = 6 & \quad 40 & 3.81 $\pm$ 0.71 & 1.89 $\pm$ 0.71 & 1.43 $\pm$ 0.71 & 1.41 $\pm$ 0.58  & & 6.14 $\pm$ 1.15 & 0.86 $\pm$ 1.11 & 0.35 $\pm$ 1.11 & 0.64 $\pm$ 0.96  \\
        b = 7 & \quad 40 &  4.84 $\pm$ 0.77 & 2.39 $\pm$ 0.73 & 1.61 $\pm$ 0.72 & 1.63 $\pm$ 0.64 & &  5.3 $\pm$ 1.17 & 1.1 $\pm$ 1.13 & 0.71 $\pm$ 1.12 & 0.82 $\pm$ 1.05  \\
                \midrule
        b = 6 & \quad 50 &  3.71 $\pm$ 0.68 & 1.54 $\pm$ 0.65 & 1.23 $\pm$ 0.64 & 1.15 $\pm$ 0.63 & &  5.3 $\pm$ 1.03 & 0.84 $\pm$ 1.0 & 0.62 $\pm$ 0.99 & 0.69 $\pm$ 0.87  \\
        b = 7 & \quad 50 &  4.01 $\pm$ 0.63 & 2.01 $\pm$ 0.66 & 1.33 $\pm$ 0.65 & 1.32 $\pm$ 0.52 & &  4.52 $\pm$ 1.12 & 0.68 $\pm$ 1.1 & 0.46 $\pm$ 1.1 & 0.43 $\pm$ 1.04  \\
        b = 8 & \quad 50 &  3.97 $\pm$ 0.7 & 2.16 $\pm$ 0.72 & 1.44 $\pm$ 0.71 & 1.53 $\pm$ 0.55 & &  4.76 $\pm$ 1.09 & 1.01 $\pm$ 1.08 & 0.54 $\pm$ 1.06 & 0.63 $\pm$ 0.95  \\
        \bottomrule
        \end{tabular}
        }
        \caption{We present average and standard error of return over 100 random seeds.}
        \label{table:appendix_main_res}
\end{table*}

\subsection{Additional Results for Our Method}\label{appendix_B3}
\begin{figure}[t]
    \centering
    \includegraphics[width=0.55\textwidth]{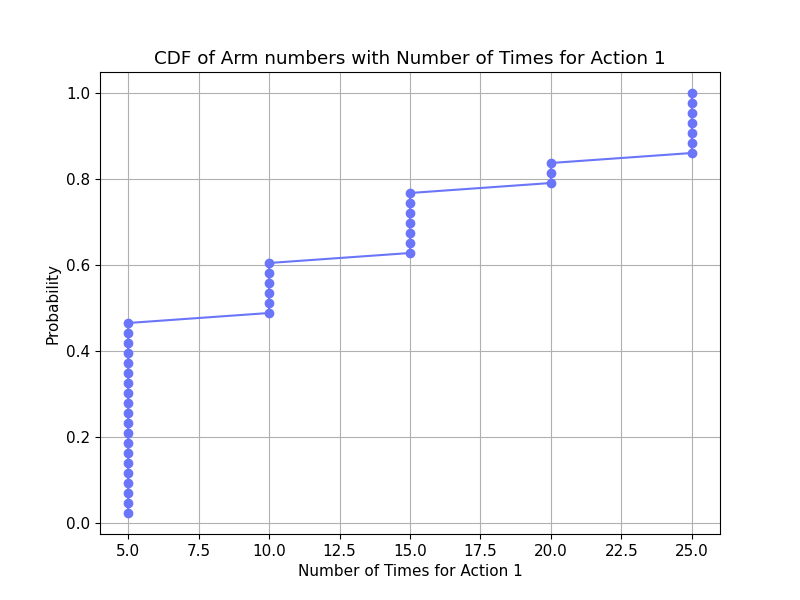}
    \caption{Cumulative Distribution Function (CDF) of the number of arms based on the number of active times (Action 1) in the MIMIC dataset. The plot shows the probability distribution of arms being active a certain number of times.}
    \label{fig:CDF_Mimic}
\end{figure}

\begin{figure*}
    \centering
    \includegraphics[width=0.8\textwidth]{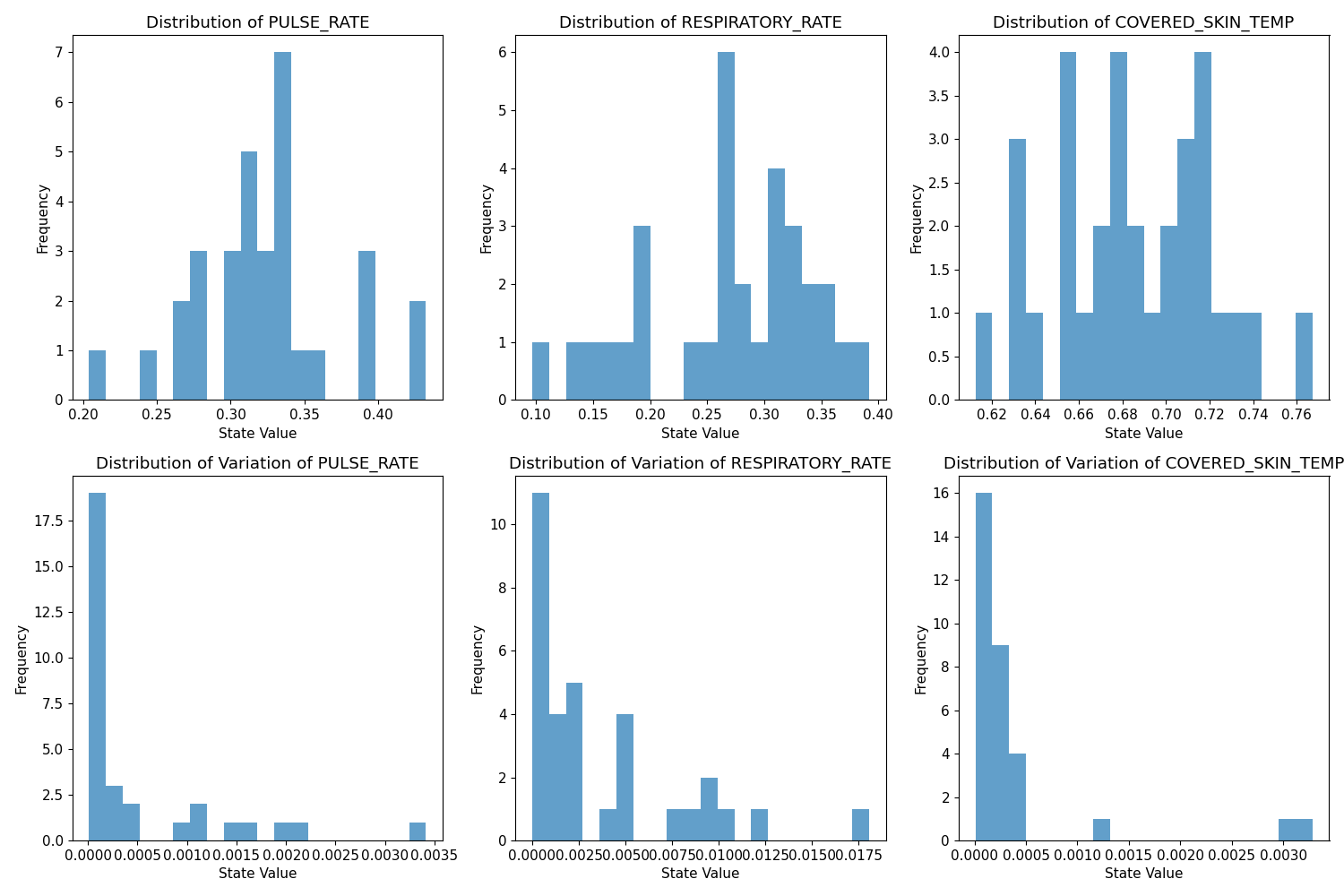}
    \caption{Analysis of the critical state dimensions that influence the decision to remove a device from a patient under the MIMIC dataset. The six state dimensions considered are PULSE\_RATE, RESPIRATORY\_RATE, COVERED\_SKIN\_TEMPERATURE, and variations of each vital sign. The histograms depict the distribution of state values before the transition from active to passive action, highlighting which factors might be most influential in triggering the change.}
    \label{fig:State_analysis_Mimic}
\end{figure*}
Figure \ref{fig:CDF_Mimic} demonstrates the Cumulative Distribution Function (CDF) of arm numbers in relation to the number of times they were active (Action 1) within the Minic dataset. The step-like nature of the CDF reflects the probability distribution across the range of active times, showing a gradual increase in cumulative probability as the number of active times grows, ultimately reaching 1.0. Notably, all arms have an active action duration larger than the minimum threshold (t\_min = 3), and 83.7\% of the arms exhibit an active action duration of less than the maximum threshold (t\_max = 25).

\begin{figure}[t]
    \centering    \includegraphics[width=0.55\textwidth]{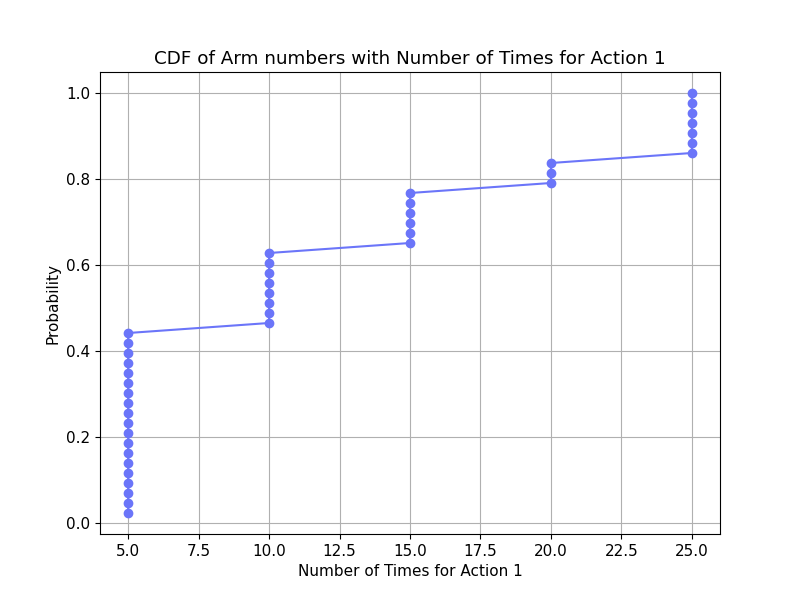}
    \caption{Cumulative Distribution Function (CDF) of the number of arms based on the number of active times (Action 1) in the Uganda dataset. The plot shows the probability distribution of arms being active a certain number of times.}
    \label{fig:CDF_Uganda}
\end{figure}

The analysis in Figure \ref{fig:State_analysis_Mimic} reveals the impact of different state dimensions on the decision to reassign devices from patients who already have them for the MIMIC dataset. Specifically, the first three dimensions, i.e., PULSE\_RATE, RESPIRATORY\_RATE, and COVERED\_SKIN\_TEMPERATURE, show that a medium value of these three vital signs significantly increases the likelihood of device reassignment.  In contrast, the last three dimensions, representing the variation in vital signs, indicate that lower variability is more likely to lead to a transition from active to passive action, thus triggering the device reassignment.

\begin{figure*}
    \centering
    \includegraphics[width=0.8\textwidth]{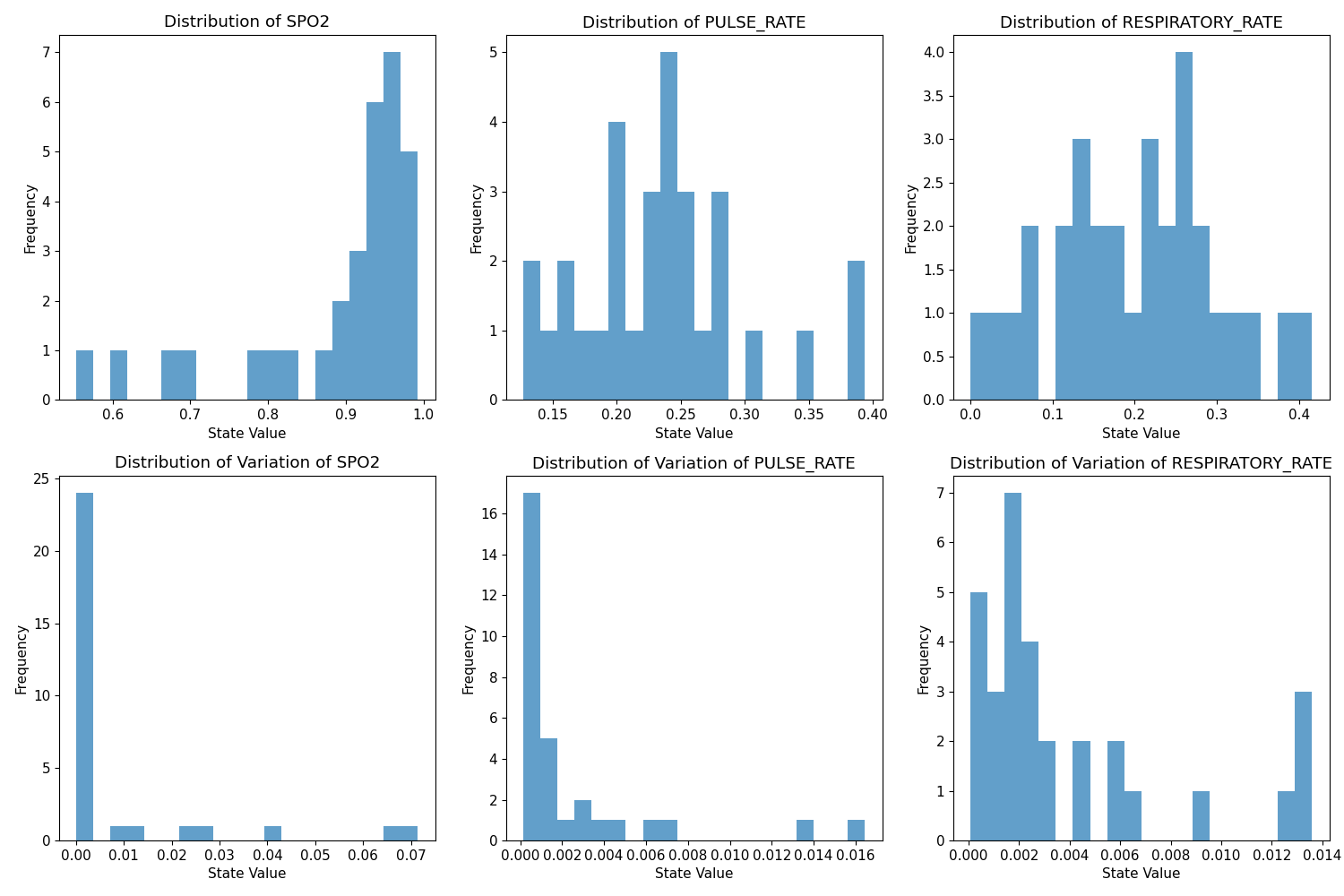}
    \caption{Analysis of the critical state dimensions that influence the decision to remove a device from a patient under the Uganda dataset. The six state dimensions considered are SPO2, PULSE\_RATE, RESPIRATORY\_RATE, and variations of each vital sign. The histograms depict the distribution of state values before the transition from active to passive action, highlighting which factors might be most influential in triggering the change.}
    \label{fig:State_analysis_Uganda}
\end{figure*}

Notice that we observe a very similar behavior of our proposed algorithm in Uganda dataset as shown in Figures \ref{fig:CDF_Uganda} and \ref{fig:State_analysis_Uganda}. Nevertheless, since Uganda data set has a different vital sign of SPO2, rather than COVERED\_SKIN\_TEMPERATURE as in MIMIC dataset, the analysis in Figure \ref{fig:State_analysis_Mimic} reveals a slightly different result. Specifically, the first dimension, SPO2, shows that a higher SPO2 value significantly increases the likelihood of device reassignment. For the second and third dimensions, PULSE\_RATE and RESPIRATORY\_RATE, medium values are more likely to trigger reassignment. Similarly, the last three dimensions, representing the variation in vital signs, indicate that lower variability is more likely to lead to a transition from active to passive action, thus triggering the device reassignment.

\begin{figure}[ht]
\centering
     \begin{minipage}
     {0.8\textwidth}
         \centering
         \includegraphics[width=0.8\textwidth]{figures/legend_cropped.pdf}
     \end{minipage}
     \begin{minipage}
     {0.8\textwidth}
         \centering
         \includegraphics[width=\textwidth]{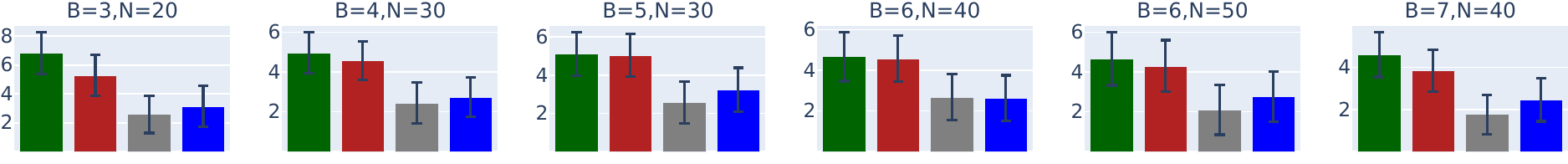}
     \end{minipage}
     \caption{Additional results on data from Mbarara, averaged over $100$ random seeds. The error bars show the standard error of rewards. Rewards are normalized by subtracting the reward of the \sname{No \ Action} baseline and then dividing by $N$.} 
        \label{fig:add_results_uganda}
\end{figure}

\end{document}